\documentclass[10pt,twocolumn,letterpaper]{article}

\usepackage{cvpr}
\usepackage{times}
\usepackage{epsfig}
\usepackage{graphicx}
\usepackage{amsmath}
\usepackage{amssymb}
\usepackage{tabularx}
\usepackage{authblk}
\usepackage{amsmath}
\usepackage{amsthm}

\usepackage[breaklinks=true,bookmarks=false]{hyperref}

\cvprfinalcopy 

\begin{document}

\title{5th Place Solution for YouTube-VOS Challenge 2022: Video Object Segmentation}
\author{Wangwang Yang*, Jinming Su*, Yiting Duan, Tingyi Guo and Junfeng Luo\\Vision Intelligence Department (VID), Meituan\\
{\tt\small \{yangwangwang, sujinming, duanyiting, guotingyi, luojunfeng\}@meituan.com}
}
\maketitle
\begin{abstract}
Video object segmentation (VOS) has made significant progress with the rise of deep learning. However, there still exist some thorny problems, for example, similar objects are easily confused and tiny objects are difficult to be found. To solve these problems and further improve the performance of VOS, we propose a simple yet effective solution for this task. In the solution, we first analyze the distribution of the Youtube-VOS dataset and supplement the dataset by introducing public static and video segmentation datasets. Then, we improve three network architectures with different characteristics and train several networks to learn the different characteristics of objects in videos. After that, we use a simple way to integrate all results to ensure that different models complement each other. Finally, subtle post-processing is carried out to ensure accurate video object segmentation with precise boundaries. Extensive experiments on Youtube-VOS dataset show that the proposed solution achieves the state-of-the-art performance with an 86.1\% overall score on the YouTube-VOS 2022 test set, which is 5th place on the video object segmentation track of the Youtube-VOS Challenge 2022.
\end{abstract}
\let\thefootnote\relax\footnotetext{* Equal contribution.}

\section{Introduction}
Video object segmentation
(VOS)~\cite{pont20172017,xu2018youtube,oh2019video,yang2020collaborative}, as a dense prediction task, aims at segmenting particular object instances across one video. Based on VOS, Semi-supervised video object segmentation (Semi-supervised VOS) targets segmenting particular object instances throughout the entire video sequence given only the object mask in the first frame, which is very challenging and has attracted lots of attention. Recently, Semi-supervised VOS has made good progress and been widely applied to autonomous driving, video editing and other fields. In this paper, we focus on improving the performance of the Semi-supervised VOS (referred to as VOS for convenience below).

In recent years, many VOS datasets have emerged, among which DAVIS\cite{pont20172017} and Youtube-VOS~\cite{xu2018youtube} are the two most widely adopted. DAVIS 2017 is a multi-object benchmark containing 120 videos with dense annotation. Compared with DAVIS, Youtube-VOS is the latest large-scale benchmark for multi-object video object segmentation and is much bigger (about 40 times) than DAVIS 2017. In Youtube-VOS, camera jitter, background clutter, occlusion and other complicated situations are kept in the process of data collection and annotation, in order to restore the real scene and solve these complicated situations by means of algorithms.
\begin{figure}[t]
\begin{center}
\includegraphics[width=1\linewidth]{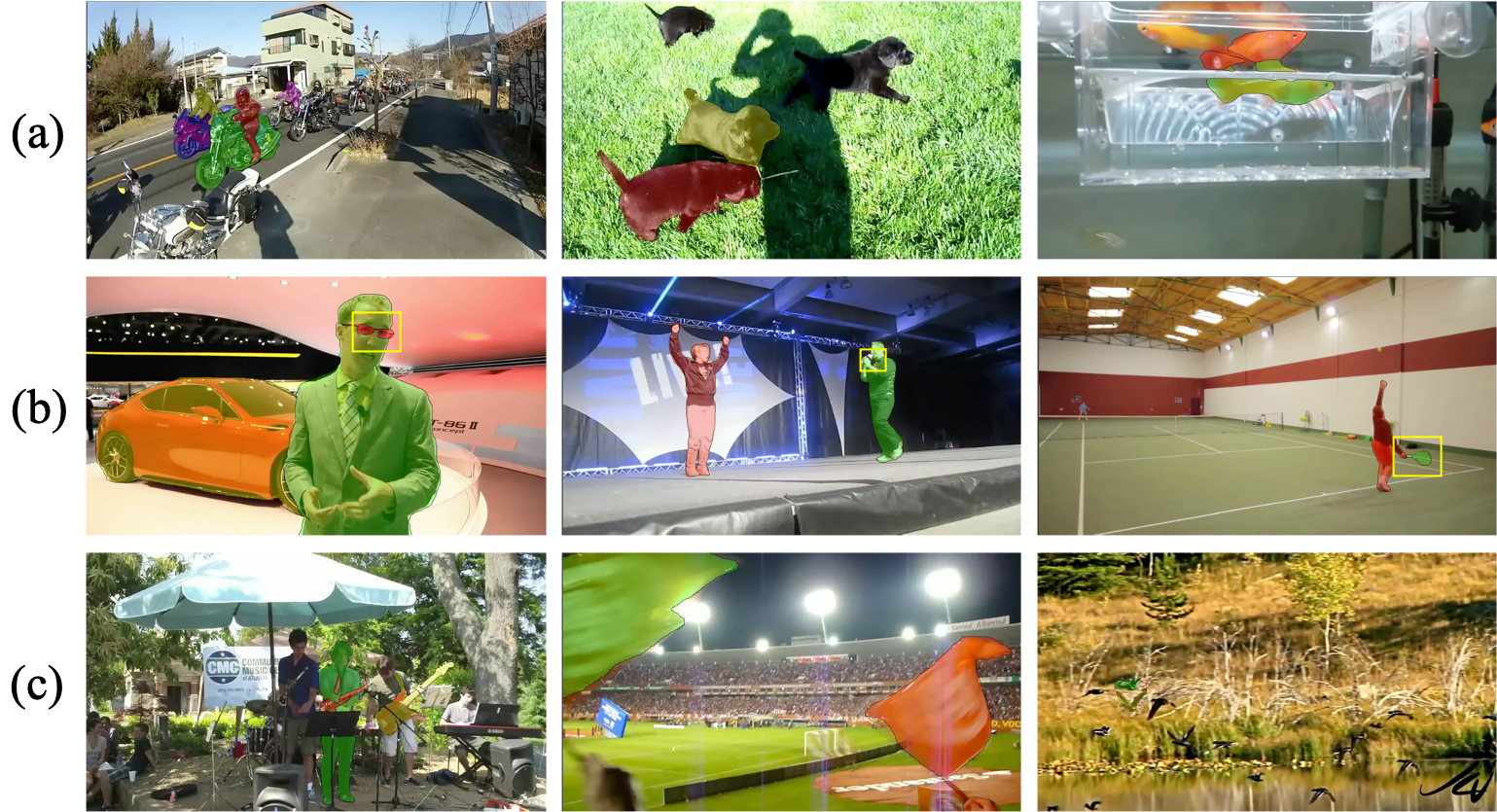}
\end{center}
   \caption{Challenges in video object segmentation. (a) Similar object are confused. (b) Tiny objects are difficult to detect. (c) Great differences in semantics and scenes bring great challenges.}
\label{fig:motivation}
\end{figure}
\begin{figure*}[t]
\begin{center}
\includegraphics[width=1.0\textwidth]{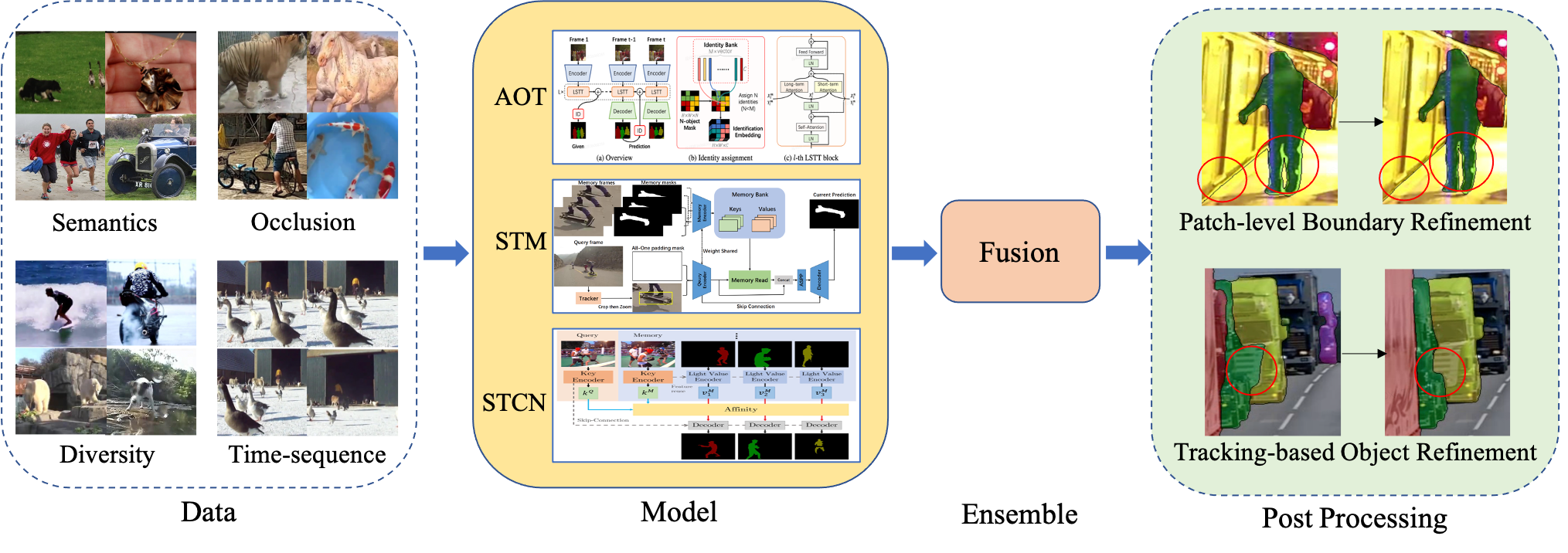}
\end{center}
   \caption{Overview of the proposed solution.}
\label{fig:framework}
\end{figure*}
To address challenges in VOS, lots of learning-based
methods have been proposed in recent years, achieving impressive performance. However, there still exist several challenges
that hinder the development of VOS. 
First of all, there are many similar objects in the real application scenarios of VOS (as shown in Figure~\ref{fig:motivation} (a)), where the accurate cross-frame tracking of these objects is very confusing, which leads to the different objects being wrongly matched as the same one.
Secondly, tiny objects are difficult to detect, especially in the process of moving across frames and the size of objects fluctuates greatly (as depicted in Figure~\ref{fig:motivation} (b)), which makes it difficult for the algorithm to accurately detect and track these objects.
In addition, many scenarios are very different, containing different objects and behaviors (as displayed in Figure~\ref{fig:motivation} (c)), which lead to many scenes and semantics not being included in the training dataset, thus bringing great challenges to the generalization of the algorithm. 
Actually, the above points are prominent problems, and there are many other difficulties to be solved in the task of VOS, which together make VOS still a challenging task.

To deal with unresolved difficulties in VOS, many methods made great efforts. Space-Time Memory network(STM)\cite{oh2019video} introduces memory networks to learn relevant information from all available sources. In STM, the past frames with object masks form an external memory, and the current frame as the query is segmented using the mask information in the memory. Specifically, the query and the memory are densely matched in the feature space, covering all the space-time pixel locations in a feed-forward fashion. By this way, it is verified that STM is able to handle the challenges such as appearance changes and occlusions effectively. In addition, Space-Time Correspondence Network  (STCN)~\cite{cheng2021rethinking} uses direct image-to-image correspondence for efficiency and more robust similarity measures in matching process, which greatly improves the efficiency and performance of STM. Recently, An associating objects with transformers algorithm (AOT)~\cite{yang2021aost} is proposed to deal with the challenging 
multi-object scenarios. In detail, AOT employs an identification mechanism to associate multiple targets into the same high-dimensional embedding space, thus simultaneously processing multiple objects' matching and segmentation decoding as efficiently as processing a single
object. Within these methods, they can track and segment most specified objects across one video, but similar objects, tiny objects and objects in complex scenes are still difficult to track and segment.

Inspired by these existing methods, we propose a simple yet effective solution for VOS, as shown in Figure~\ref{fig:framework}. In order to deal with existing difficulties, we first analyze the Youtube-VOS dataset and other video segmentation related datasets (\eg, OVIS\cite{qi2022occluded} and VSPW\cite{miao2021vspw}). We find that other datasets can supplement the diversified scenes with similar objects and occlusion situation from the data aspect. And in the model aspect, we choose three different basic networks (\ie, AOT\cite{yang2021aot}, STCN\cite{cheng2021rethinking} and FAMNet\cite{yangfeature}(An improved STM\cite{oh2019video})), which have different structures and can learn the information of objects from different aspects, so as to achieve complementary promotion. Next, a simple fusion method is used to integrate different predictions from several variants based on the above three basics. Finally, a series of exquisite post-processing strategies are introduced to improve the prediction results and ensure accurate video object segmentation with precise boundaries. Extensive experiments on the Youtube-VOS dataset show that the proposed solution achieves state-of-the-art performance.

The main contributions of this paper include: 1) We analyze the characteristics of the Youtube-VOS dataset, and supplement the dataset with static and video segmentation datasets. 2) We improve three basic network architectures and train several variants to learn the different aspects of information for objects in videos. 3) We introduce a series of subtle fusion and post-processing strategies to ensure accurate video object segmentation with precise boundaries. 4) The proposed solution achieves the state-of-the-art performance with an 86.1\% overall score on the YouTube-VOS  2022 test set, which is the 5th on the video object segmentation track of the Youtube-VOS Challenge 2022.

\section{Method}
To address difficulties in VOS, we propose a simple yet effective solution, as shown in Figure~\ref{fig:framework}. Details of the proposed solution are described as follows.

\subsection{Data Matters}
Learning-based VOS methods are highly dependent on data. Although YouTube-VOS is the largest video object segmentation dataset, it is still unable to utilize the potential of the current state-of-the-art methods sufficiently. 

Through the analysis of the Youtube-VOS dataset, we find that there are four key points to pay attention to at the data level, which are summarized as semantics, occlusions, diversity and time-sequence. Most state-of-the-art methods in the VOS task adopt a two-stage training strategy. In the first stage, video clips synthesized from the static images are used for pre-training. Then the real video data is used for final training in the second stage. Large-scale static image datasets come from fields like instance segmentation and salient object detection, and therefore have more semantics and diversity. By pre-training on them, the VOS model can extract robust feature embedding for pixel-level spatiotemporal feature matching, and improve the ability to identify and discriminate against diverse targets. We try to introduce ImageNet-Pixel\cite{zhang2020interactive}, a more diverse image dataset in the pre-training stage, but it does not bring obvious benefits. We believe that this is because the current model structure and separated two-stage training method cannot fully utilize the information in the static image datasets. On the other hand, video data in the real world have additional temporal information compared to static image data, and the data form in the testing stage is video, so it is more straightforward to bring more video segmentation datasets to improve performance. Benefit from the release of several new datasets in the video segmentation field recently, such as YoutubeVIS which has more objects in each video, OVIS~\cite{qi2022occluded} which occlusion scenarios are significant, and VSPW~\cite{miao2021vspw}) which have dense annotations and high-quality resolution, we introduce them into the second training stage, thus significantly improving the performance of models.

After the above data supplement, the ability of the model in the aspect of semantics extraction, occlusions recognition and other aspects has been enhanced.

\subsection{Strong Baseline Models}
We adopt three kinds of architectures as our baseline models including AOT\cite{yang2021aot}, FAMNet\cite{yangfeature}, and STCN\cite{cheng2021rethinking}, which have high performance in the VOS field recently. The detailed implementation can be found in their original papers.

Benefit from an effective identification mechanism and long short-term transformer module, AOT can achieve high performance for both seen and unseen objects in the test phase. Meanwhile, FAMNet and STCN can produce better results for unseen objects because of the simplicity and robustness of their core pixel-wise feature matching module. By combining methods with different network designs, we can get several sets of results at different aspects of information for video objects and obtain more gains in the model ensemble stage.

\subsection{Nontrivial Attempts}
\noindent\textbf{LSTT block V2:} Although the network structure of the AOT\cite{yang2021aot} is delicate, it still has room for improvement. Following AOST\cite{yang2021aost}, we improve Long Short-Term Transformer (LSTT) block, the core module in AOT model, and obtain the LSTT block V2 which has better performance. Specifically, LSTT block utilize the attention mechanism to perform pixel-level feature matching between the current frame and memory frames. The formulas of common attention-based matching mechanism, attention-based matching mechanism of the LSTT block and attention-based matching mechanism of the LSTT block V2 are,
\begin{equation}
\begin{aligned}
    \label{eqn:attention}
    Att(Q,K,V) &=Corr(Q,K)V\\&= softmax(\frac{QK^{tr}}{\sqrt{C}})V
\end{aligned},
\end{equation}
\begin{equation}
    \label{eqn:aot attention}
    Att(Q,K,V+E),
\end{equation}
\begin{equation}
    \label{eqn:aost attention}
    Att(Q,K\odot\sigma(W_{l}^{G}E),V+W_{l}^{ID}E).
\end{equation}
By combining the target identification embedding E with the value embedding of the memory frames V in Eq. \ref{eqn:attention}, AOT can propagate the information of multiple targets to the current frame simultaneously (Eq. \ref{eqn:aot attention}).  Compared with the original LSTT block, the LSTT block V2 is more effective. There are two main differences between them. The first one is in the value part of the attention mechanism. LSTT block V2 projects E by a linear layer $W_{l}$ whose weights are various in different LSTT layers. Such modification increases the degree of freedom of the model. The second one is in the key part of the attention mechanism. LSTT block V2 generates a single channel map to adjust the key embedding of the memory frames K using E so that the target information of memory frames can be used in the matching process between Q and K too.
\\
\noindent\textbf{Turning off strong augmentation:} In order to further improve the performance of our models, a trick frequently used in the object detection field is performed. Specifically, we turn off data augmentation operations other than random cropping for the last few epochs when running the second stage of training. Meanwhile, we only use YouTube-VOS as our training data. In this way, the data distribution in the final training stage is more consistent with the data distribution in the testing stage.
\\
\noindent\textbf{Attaching top-k filtering to memory read:} The long-term memory bank in AOT is similar to the memory bank in STM-like methods. So the number of memory frames used in the training stage and testing stage is inconsistent. Besides, as the video's length increases in the testing stage, the size of long-term memory bank also grows dynamically. So we attempt to add top-k filtering operation\cite{cheng2021mivos} to the Long-Term Attention module in AOT, to alleviate the problem of information redundancy and remove noises in long-term memory. But this attempt doesn't always work in all of our models.
\\
\noindent\textbf{Model ensemble:} Considering that the performance of different models is diverse, we adopt offline ensembling to fuse these models' predictions for getting higher precision frame by frame. Specifically, we have tried two fusion methods. First, we average predictions of all models directly, to help the models complement each other and reduce error prediction. The second interesting idea is keypoint voting, we use feature matching\cite{sp,sg}to correlate the target in the previous frames and current frame, so as to judge the quality of the prediction of different models in the current frame and weight them by keypoint voting,  which reduces some wrong predictions.
\\
\noindent\textbf{Patch-level boundary refinement and tracking-based small object refinement:} In addition to the above attempts, we use some post-processing strategies to refine the predicted results. We adopt boundary patch refinement (BPR)\cite{bpr} to improve the boundary quality of object segmentation. BPR is a conceptually simple yet effective post-processing refinement framework. After that our predictions have significant improvements near object boundaries, as shown in Figure~\ref{fig:framework}. Besides, the input of most existing state-of-the-art methods is the whole video frame, and the resolution of the feature map in the pixel-level feature matching process is further reduced because of the downsampling operation. Both of them cause poor results for small objects. Therefore, we adopt the crop-then-zoom strategy in FAMNet. Firstly, we integrated box-level object position information provided by the tracker\cite{TransT} and the preliminary segmentation result of the object provided by the VOS model to get the approximate positions of partial small objects within the dataset in every frame. Then the original frames are cropped and resized to a larger size. Finally, a secondary segmentation is performed on the clip to obtain more accurate segmentation results for small objects.

\section{Experiments}
\subsection{Training Details}
To comprehensively improve the accuracy, four different frameworks, including AOT\cite{yang2021aot}, improved AOT\cite{yang2021aost}, STCN\cite{cheng2021rethinking}, and FAMNet\cite{yangfeature} are used in our experiments. For AOT, multiple networks such as Swin\cite{swin}, EfficientNet\cite{efficientnet}, and ResNext\cite{resnext} are used as the encoder to obtain better accuracy. Noted that the parameters of BN layers and the first two blocks in the encoder are frozen in view of stabilizing the training. Following the official settings of AOT, we also take a two-stage training strategy. In the pre-training stage, several static image datasets including COCO\cite{coco}, ECSSD\cite{ecssd}, MSRA10K\cite{MSRA10K}, PASCAL-S\cite{pascal-s}, PASCAL-VOC\cite{pascal-voc} are used for preliminarily semantic learning. During the main training, video datasets including Youtube-VOS\cite{xu2018youtube}, DAVIS 2017\cite{pont20172017}, YouTubeVIS\cite{youtubevis}, OVIS\cite{qi2022occluded}, and VSPW\cite{miao2021vspw} are used to enhance the generalization and robustness of the model.

During the training of AOT, images are cropped into patches with the fixed size of 465 $\times$ 465, and multiple image and video augmentations are randomly applied following \cite{RGMP,CFBI} to enrich the diversity of data. The experiments are performed on Pytorch by using 4 Tesla A100 GPUs. We minimize the sum of bootstrapped cross-entropy loss and soft Jaccard loss by adopting AdamW\cite{adamw} optimizer. In the pre-training, the initial learning rate is set to 4e-4 with a weight decay of 0.03 for 100,000 steps, and the batch size is set to 32 for acceleration. In the main training, with the initial learning rate of 2e-4, weight decay of 0.07, and batch size of 16, the training step is extended to 130,000 due to the expansion of data. Noted that all learning rates will decay to 2e-5 by using a polynomial manner as \cite{CFBI}. To ensure the stability of training and enhance the robustness of the model, we also adopt the Exponential Moving Average (EMA)\cite{ema} to average the parameters of the model for better performance. For the training of STCN and FAMNet, all training process follows their official implementations.
\subsection{Inference and Evaluation}
For evaluating the single model of AOT, to reduce the sensitivity of the model to various object scales, online flip and multi-scale testing is applied to obtain better accuracy. Specifically, the predictions generated from videos with three scales of 1.2 × 480p resolution, 1.3 × 480p resolution, and 1.4 × 480p resolution are ensembled frame by frame during the inference. In addition, to balance the number of objects that appear in the quite long or quite short video sequence, a dynamic memory frame sampling strategy is introduced in our experiments. For STCN and FAMNet, we also apply the flip and multi-scale testing in our experiments.
\begin{figure*}[ht]
\begin{center}
\includegraphics[width=1.0\textwidth,height=7.5cm]{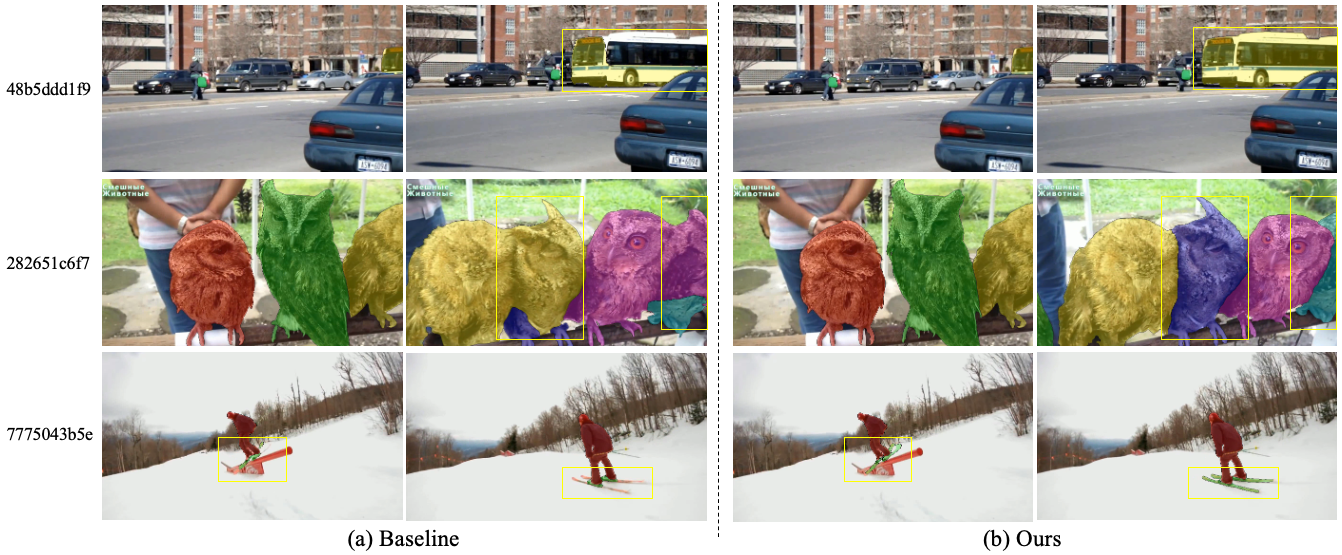}
\end{center}
   \caption{Representative visual examples from the baseline model and the proposed solution.}
\label{fig:results}
\end{figure*}
\begin{table*}[t]
\centering
\renewcommand\arraystretch{1.1}
\setlength{\tabcolsep}{6mm}{
\begin{tabular}{c | c | c | c | c | c}
\hline
\textbf{Team Name} & \textbf{Overall} & \textbf{$\mathcal{J}_{seen}$} & \textbf{$\mathcal{J}_{unseen}$} & \textbf{$\mathcal{F}_{seen}$}  & \textbf{$\mathcal{F}_{unseen}$}\\
\hline
Thursday\_Group & 0.872(1) & 0.855(1) & 0.817(3) & 0.914(1) & 0.903(1) \\
ux & 0.867(2) & 0.844(3) & 0.819(1) & 0.903(2) & 0.903(2) \\
zjmagicworld & 0.862(3) & 0.841(4) & 0.816(4) & 0.895(4) & 0.896(4)\\
whc & 0.862(4) & 0.840(5) & 0.818(2) & 0.894(5) & 0.896(5)\\
\textcolor[rgb]{0.00,0.00,1.00}{gogo} & \textcolor[rgb]{0.00,0.00,1.00}{0.861(5)} & \textcolor[rgb]{0.00,0.00,1.00}{0.847(2)} & \textcolor[rgb]{0.00,0.00,1.00}{0.808(7)} & \textcolor[rgb]{0.00,0.00,1.00}{0.901(3)} & \textcolor[rgb]{0.00,0.00,1.00}{0.890(6)}\\
sz & 0.857(6) & 0.831(6) & 0.815(5) & 0.886(7) & 0.896(3)\\
PinxueGuo & 0.856(7) & 0.832(7) & 0.812(6) & 0.887(6) & 0.892(7)\\
\hline
\end{tabular}}
\vspace{0.3cm}
\caption{Ranking results in the YouTube-VOS 2022 test set. We mark our results in blue.}
\label{table:ranking}
\end{table*}

Considering that models with different structures have unique predictive advantages, we adopt an offline model ensemble strategy to further improve the performance of results. Specifically, soft prediction scores produced by models which have different frameworks and different backbone networks are simply averaged as the final result. Noted that we also have tried other strategies like max weighting and key-point voting, and the average operation gains the best performance. All the results are evaluated on the official YouTube-VOS evaluation servers.

\subsection{Results}

Through test-time augmentation, model ensemble and post-processing strategies, the proposed solution obtain the 5th place on the YouTube-VOS 2022 testset, as listed in Table~\ref{table:ranking}. From the result, we see that our solution surpasses most solutions in the seen category (as shown in $\mathcal{J}_{seen}$ and $\mathcal{F}_{seen}$), which is a characteristic of our solution. In addition, we also show some of our quantitative results in Figure~\ref{fig:results}. It can be seen that the proposed solution can accurately segment objects in some difficult scenarios which have severe changes in object appearance, confusion of multiple similar objects and small objects.

In order to demonstrate the effectiveness of different components, we conduct several ablation experiments. Quantitative results are shown in Table~\ref{table:ablation}. We boost the performance of the original AOT network to 86.6\% on YouTube-VOS 2019 validation set without any test-time augmentation such as multi-scale testing or flip testing.

\begin{table}[h]
\centering
\begin{tabular}{p{50mm}<{\centering}|p{15mm}<{\centering}}
\hline
\textbf{Components} & \textbf{Overall} \\
\hline
AOTL-R50(baseline) & 85.3\\
+ Swinb backbone & 85.5\\
+ LSTT Block v2 & 85.7\\
+ More real video data & 86.2\\
+ Turn off strong augmentation  & 86.6\\
\hline
\end{tabular}
\vspace{0.3cm}
\caption{Ablation study on YouTube-VOS 2019 validation set.}
\label{table:ablation}
\end{table}
\section{Conclusion}
In this paper, we propose a solution for the video object segmentation task, and make nontrivial improvements and attempts in many stages such as data, model, ensemble, and post-processing strategies. In the end, we achieve the 5th place on the YouTube-VOS 2022 Video Object Segmentation Challenge with an overall score of 86.1\%.
{\small
\bibliographystyle{ieee}
\bibliography{egbib}
}
\end{document}